\documentclass[twoside,11pt]{article}

\usepackage{amsmath}

\usepackage[accepted]{ewrl_2022}

\usepackage{lipsum}  
\usepackage{todonotes}

\usepackage{caption}
\usepackage{subcaption}

\usepackage{xspace}

\usepackage{caption}
\usepackage{graphicx}
\usepackage{subcaption}

\usepackage{algorithm}
\usepackage{algorithmic}

\usepackage{xcolor} 
\usepackage{tikz}

\usetikzlibrary{patterns}
\usetikzlibrary{calc}
\usetikzlibrary{decorations.pathmorphing} 
\usetikzlibrary{fit}					
\usetikzlibrary{backgrounds}	

\usetikzlibrary{shapes.geometric}
\usetikzlibrary{arrows,shapes}
\usetikzlibrary{trees,snakes}
\usetikzlibrary{decorations.markings}
\usetikzlibrary{shadows}
\usetikzlibrary{positioning}

\ewrlheading{15}{2022}{September 2022, Milan, Italy}{Riccardo Della Vecchia, Alena Shilova, Philippe Preux, Riad Akrour}

\newcommand{\politex}{\textsc{Politex}\xspace}
\newcommand{\ourAlgo}{\textsc{MicaRL}\xspace} 

\newcommand{\RR}{\mathbb{R}}
\newcommand{\E}{\mathbb{E}}

\newcommand{\policy}[1][]{\pi_{#1}}
\newcommand{\approxQ}[1][]{\hat Q_{#1}}

\newcommand{\Q}[1][]{Q_{#1}}
\newcommand{\V}[1][]{V_{#1}}
\newcommand{\env}{\mathcal{S}}
\newcommand{\act}{\mathcal{A}}

\newcommand{\data}[1][]{\mathcal{D}^{#1}}
\newcommand{\feat}[1][]{\phi^{(#1)}}
\newcommand{\prefeat}[1][]{\tilde{\phi}^{(#1)}_{\xi^{(#1)}}}
\newcommand{\outW}[1][]{W^{(#1)}}
\newcommand{\qW}[1][]{W_Q^{(#1)}}
\newcommand{\deltaW}[1][]{W_{\delta}^{(#1)}}

\newcommand{\nbiter}{\text{NB\_ITER}}
\newcommand{\nbsamp}{\text{NB\_SAMP}}
\newcommand{\bsize}{\text{batch-size}}
\newcommand{\concat}{\textsc{cat}}
\newcommand{\rlberry}{\texttt{rlberry}}

\ShortHeadings{Mirror Reinforcement Learning with Cascade Neural Networks}{Della Vecchia, Shilova, Preux, Akrour}
\firstpageno{1}

\begin{document}

\title{
Entropy Regularized Reinforcement Learning with Cascading Networks}

\author{\name Riccardo Della Vecchia$^*$ \email riccardo.della-vecchia@inria.fr \\
       \addr Univ. Lille, Inria, CNRS, Centrale Lille, UMR 9189 - CRIStAL \\
       Lille, France
      \AND
      \name Alena Shilova$^*$ \email alena.shilova@inria.fr \\
      \addr Univ. Lille, Inria, CNRS, Centrale Lille, UMR 9189 - CRIStAL\\
      Lille, France
      \AND
        \name Philippe Preux \email philippe.preux@inria.fr \\
      \addr Inria, Univ. Lille, CNRS\\
      Lille, France
      \AND
        \name Riad Akrour$^*$ \email riad.akrour@inria.fr \\
      \addr Univ. Lille, Inria, CNRS, Centrale Lille, UMR 9189
- CRIStAL\\
      Lille, France
    }
\maketitle
\def\thefootnote{*}\footnotetext{These authors contributed equally to this work.}
\def\thefootnote{\arabic{footnote}}

\begin{abstract}
Deep Reinforcement Learning (Deep RL) has had incredible achievements on high dimensional problems, yet its learning process remains unstable even on the simplest tasks. Deep RL uses neural networks as function approximators. These neural models are largely inspired by developments in the (un)supervised machine learning community. Compared to these learning frameworks, one of the major difficulties of RL is the absence of i.i.d. data. One way to cope with this difficulty is to control the rate of change of the policy at every iteration. In this work, we challenge the common practices of the (un)supervised learning community of using a fixed neural architecture, by having a neural model that grows in size at each policy update. This allows a closed form entropy regularized policy update, which leads to a better control of the rate of change of the policy at each iteration and help cope with the non i.i.d. nature of RL. Initial experiments on classical RL benchmarks show promising results with remarkable convergence on some RL tasks when compared to other deep RL baselines, while exhibiting limitations on others.
\end{abstract}


\section{Introduction}
\label{sec.intro}
Reinforcement learning (RL) formulates a general machine learning problem in which an agent has to take a sequence of decisions to maximize a supervision signal~\citep{Sutton18, Szepesvari10}. RL as a learning framework, especially when combined with neural function approximators, has made large practical breakthroughs over the last years on complex, high dimensional problems \citep{Go,Atari}, and its range of application continues to grow to new domains such as drug discovery \citep{FlowRL}. Surprisingly however, even simple RL tasks can exhibit one of the main disadvantages of current deep RL algorithms: their training is unstable and does not exhibit convergence to an optimal policy. Rather, deep RL discovers good policies along the way with often large performance oscillations in-between.

So what makes the training process of a neural model by RL more brittle than say, the training of the same model to regress a continuous signal or classify images? Certainly, one of the major challenges of RL, and specifically online RL\footnote{We note that offline, batch, RL also violates the independent and identically distributed assumption since there is still a mismatch between the distribution of training data and the data generated by the learned policy, which is inherent to the sequential nature of RL in general. However, this work will only focus on the online RL setting.}, is that the agent's decisions influence the data gathering process, violating the typical assumption of the aforementioned supervised learning tasks that data is independent and identically distributed~\citep{bishop2006}. To cope with this challenge, several RL algorithms constrain the agent's behavior to only slowly change. In trajectory optimization and optimal control, a new policy is made close to the policy around which the dynamics of the system have been approximated through mixing \citep{Todorov2005, Tassa14} or by a Kullback-Leibler (KL) constraint \citep{Levine2014}. In policy gradient, a key breakthrough was the use of natural gradient that follows the steepest descent in policy space rather than parameter space \citep{bagnell2003, Peters08, Bhatnagar09}, i.e. seeking maximal objective improvement with minimal \textit{policy} change. 

Constraining successive policies to be close to each others in approximate policy iteration is justified in the seminal work of \citealt{Kakade02} by the mismatch between what the policy update should optimize (advantage function in expectation of its own state distribution) and what is optimized in practice (advantage function in expectation of the state distribution of the data gathering policy). As in optimal control, closeness can be achieved by mixing policies~\citep{Kakade02, Pirotta2013a}, limiting deviation of their probability ratio to one \citep{Schulman17} or constraining their KL-divergence \citep{Schulman15, Abdolmaleki18, Tangkaratt18, Akrour18}. In the latter case, when KL-divergence is used to regularize the policy update, a recent line of research has drawn similarities between these algorithms and the convex optimizer mirror descent \citep{Neu17, Geist19, politex}, which provides as a result convergence guaranties to an optimal policy. Of course, practical implementations of these algorithms may still fail to find the optimal policy due to two difficulties: \textbf{i)} learning the Q-function of the current policy from sample data and \textbf{ii)} solving the entropy regularized policy update in the neural parameter space.

In this paper, we investigate the use of Cascade Neural Networks~(Cascade-NN, \citep{cascadeNN}) in the context of entropy regularized policy iteration to address \textbf{ii}. Unlike typical neural models, Cascade-NNs can grow in size during learning. Importantly, when new neurones are added, old ones are frozen which allows us to perform the policy update in closed form and completely eliminate the second source of errors discussed above. In the remainder of the paper, we will first describe in a preliminaries section Cascade-NNs and entropy regularized RL (Section~\ref{sec.prelim}), before describing our approach in Section~\ref{sec.method}. We discuss related work in Section~\ref{sec.rel} after comparing our algorithm in Section~\ref{sec.exp} to deep RL baselines on classical RL benchmarks.

\section{Preliminaries}
\label{sec.prelim}
In this section, we introduce our notations and briefly describe the framework of entropy regularized RL. We then present the general architecture of Cascade-NNs as described in the seminal paper of \citep{cascadeNN} before describing how it is used in our algorithm. 

\subsection{Notation}
\label{sec.notation}

We consider Markov Decision Problems (MDPs) defined as a tuple $(\env, \act, r, P, \gamma)$ to model the interactions of the agent with the environment. $\env$ is a finite set of states, $\act$ is a finite set of actions, $r$ is the unknown reward function 
$r: \env \times \act \rightarrow \RR$ and $P$ is the unknown probability transition 
function $P: \env \times \act \rightarrow \Delta_{\env}$, where $\Delta_{\env}$ is the set of distributions over ${\env}$. 

Let the policy $\policy: \env \rightarrow \Delta_{\act}$ be a mapping between a state to a distribution over actions. For every such policy $\policy$, one can compute the value function
$$\V[\policy] (s) = \E \Biggl[ \sum_{t=0}^{\infty} \gamma^{t} r(s_t, a_t) \Biggm| s_0 = s \Biggr],$$
where the expectation is taken w.r.t. all states and actions following $s$.
We also define the state-action value function as 
$$\Q[\policy](s,a) = r(s,a) + \gamma \E_{s'\sim P(s,a)} \V[\policy] (s').$$



\subsection{Entropy regularized policy iteration}
\label{sec.entreg}
RL considers the problem of finding the optimal policy in the MDP's unknown environment. One way of doing so is to use policy iteration methods that successively perform \textbf{i)} a policy evaluation step to compute $\Q[\policy]$ and \textbf{ii)} a policy improvement step, yielding a new policy $\policy'$ that picks at every state $s$ an action in $\arg\max_a \Q[\policy](s, a)$. Repeatedly performing these two steps will converge in finite time to an optimal policy, irrespective of the choice of the initial policy~\citep{Sutton18}. However, we notice that at any iteration, the policy $\policy'$ obtained after a policy improvement step does not explore actions that do not maximize the previous Q-function $\Q[\policy](s, .)$, for every state $s$. Hence, to correctly estimate $\Q[\policy'](s, .)$ for such actions in practice, one would need to introduce another data gathering policy that is more explorative than $\policy'$. An alternative is to regularize the policy update step to maintain the stochasticity of $\policy'$. This is usually performed by adding a KL-divergence term between $\policy$ and $\policy'$, ensuring that exploration does not vanish too quickly. 

In contrast to these algorithms, in this paper, we investigate solving the soft policy update exactly for every state, using an incrementally increasing neural architecture. Given $\approxQ[{\policy[i]}]$, the approximation of the Q-function of $\policy[i]$ at iteration $i$ of the algorithm, our policy update at every state $s$ is similar to that of MPO \citep{Abdolmaleki18} and \politex \citep{politex} and is given by the following optimization problem:
\begin{equation}
    \label{eq.mirror}
    \begin{split}
        \policy[i+1](s) &= \arg \max_{\policy} \E_{a \sim \policy(.|s)}\approxQ[{\policy[i]}](s, a) 
        - \frac1{\eta} \text{KL}(\policy(s) \| \policy[i](s)), \\
        & \propto \exp \left( \eta \approxQ[{\policy[i]}] (s, \cdot) \right)\policy[i](.|s),
    \end{split}
\end{equation}
where $\text{KL}(\policy(s) \| \policy[i](s))$ is the KL-divergence between the distributions over $\act$ that are $\policy(s)$ and $\policy[i](s)$ at state $s$. By recursion and assuming that $\policy[0]$ is the uniform distribution over $\act$, we get
\begin{align}
    \policy[i+1](s) \propto \exp \left( \eta \sum_{k=0}^i \approxQ[{\policy[k]}] (s, \cdot) \right).
    \label{eq.policy}
\end{align}
We thus see that the entropy regularized policy update can be solved in closed form, and policies have a very simple form provided we can keep track of all previously estimated Q-functions. Of course, this might appear too cumbersome in practice if we consider that each Q-function is a separate neural network. For this reason, other papers approximated the update, e.g. in MPO~\citep{Abdolmaleki18}, by fitting a Gaussian neural policy minimizing the divergence to $\exp \left( \eta \approxQ[{\policy[i]}] (s, \cdot) \right)\policy[i](.|s)$. Alternatively, in \citep{improvedpolitex}, authors have considered keeping only the last few Q-functions or keeping a large replay memory of past MDP transitions, and fitting a single network to the sum of Q-functions. In contrast, our policy will take the exact form of Eq.~\eqref{eq.policy}. However, we will not train independent neural networks for each $\approxQ[{\policy[k]}]$, but leverage the Cascade-NN architecture and only add a few neurons at every iteration. The rationale is that since the policy will not change drastically between iterations neither will their Q-function and one might need only a few more neurons to learn $\delta_k = \approxQ[{\policy[k]}] - \approxQ[{\policy[k-1]}]$.

\subsection{Cascade-NN}
\label{sec.cascade}

\begin{figure}
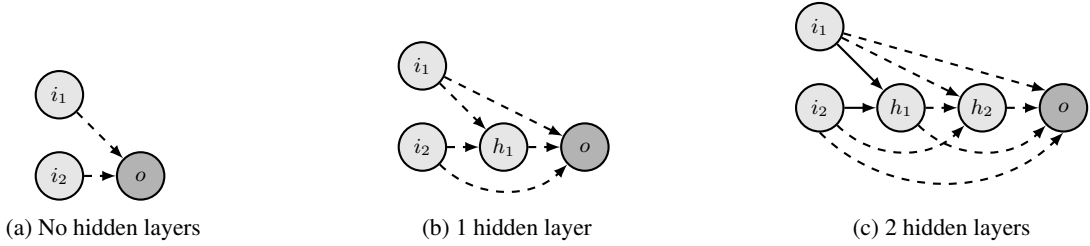

    \centering
    \begin{subfigure}[b][][t]{0.3\textwidth}
        \centering
        \input{fig/diagrams/cascade_tikz-1}
        \caption{No hidden layers}
        \label{fig.cascade1}
    \end{subfigure}
    \hfill
    \begin{subfigure}[b][][t]{0.3\textwidth}
        \centering
        \tikzstyle{output} = [circle, thick, minimum size = 0.7cm, draw = black, fill = black!30, 
                              inner sep=0pt, font = \small, node distance = 1.2cm]
\tikzstyle{other} = [circle, thick, minimum size = 0.7cm, draw = black, fill = black!10, 
                              inner sep = 0pt, font = \small, node distance = 1.2cm]
\tikzstyle{Tedge} = [->, dashed]
\tikzstyle{TCedge} = [->, dashed, bend right = 45]
\tikzstyle{Fedge} = [->, solid]
\tikzstyle{FCedge} = [->, solid, bend right = 45]

\begin{tikzpicture}[>=latex, thick, scale=1, every node/.style={scale=0.9}] \tiny
\node (i1)[other]{$i_{1}$};
\node(i2)[other, below of=i1]{$i_{2}$};
\node(h1)[other, right of=i2]{$h_{1}$};
\node(out)[output, right of =h1]{$o$};
\draw[Tedge] (i1) -- (out);
\draw[Tedge] (i2) -- (h1);
\draw[Tedge] (i1) -- (h1);
\draw[Tedge] (h1) -- (out);
\path (i2) edge[TCedge] (out);
\end{tikzpicture}
        \caption{1 hidden layer}
        \label{fig.cascade2}
    \end{subfigure}
    \hfill
    \begin{subfigure}[b][][t]{0.35\textwidth}
        \centering
        \input{fig/diagrams/cascade_tikz-3}
        \caption{2 hidden layers}
        \label{fig.cascade3}
    \end{subfigure}
    \caption{Incremental growth in Cascade-NN architecture and freezing of weights. Dashed lines show learnable parameters after each addition, while solid lines represent frozen weights. In a Cascade-NN, when new hidden nodes are added, all inputs to older nodes are frozen so as to freeze older features.}
    \label{fig:Cascade}
\end{figure}

Cascade Neural Network (Cascade-NN) or also called Cascade-Correlation 
Networks~\citep{cascadeNN} is a special type of neural network architecture that is growing at each epoch by $n$ (typically $n=1$) new neurons that are connected to the input of the NN and all previously created hidden neurons. At first, it has no
hidden layer (Figure~\ref{fig.cascade1}), and gradually more and more
neurons get added to its structure (Figures~\ref{fig.cascade2} and~\ref{fig.cascade3}). 
Historically, these newly added neurons were trained following a three phase process. Firstly, a batch of neurons larger then $n$ is added and the output of each of those new neurons is trained to maximize the correlation
with the current residual error of the model. 
Secondly, the neurons are ranked according to their correlation and only the $n$-top neurons are kept for the final phase. Thirdly, new neurons are connected to the output layer and the weights of the output layer are updated in order to minimize the residual error. Importantly, throughout all phases older neurons are frozen, which means that one can easily compute the sum of all past outputs $\sum_k o_k$ by simply summing all past weights of the (linear) output layer. An illustration of the freezing procedure is given in Figure~\ref{fig:Cascade}, showing all re-trainable parameters at every iteration. Here $i_1$ and $i_2$ correspond to data inputs, $h_1$ and $h_2$ to two hidden neurons and $o$ to the output. Solid edges show the frozen connections, while dashed edges are for trainable parameters.

In~\citep{cascadeNN}, the authors emphasize the following advantages of Cascade-NN with respect to classical MLP neural networks:
\begin{itemize}
    \item \textbf{Non-parametric training}, there might be less hyper-parameters to tune such as the depth, width and connectivity of NN, its training (and growth) is stopped automatically as soon as a stopping criterion is satisfied;
    \item \textbf{Fast learning}, freezing all the layers except the last one helps to optimize the parameters without doing back-propagation, in this way all neurons have their independent goal and they can "settle into distinct useful roles";
    \item \textbf{Incremental learning}, especially useful when the model constantly receives  some new information (data) in a stream manner, in this case old features are preserved, while new features enrich the feature extraction for the newly obtained data.  
\end{itemize}

Prior work already explored the use of Cascade-NN in RL. Notably, \citep{PhilippeCascade} combined features trained to maximize correlation with the Bellman residual before using LSPI~\citep{lspi} to find an optimal policy given the current set of features. However, \citep{PhilippeCascade} did not leverage the properties of Cascade-NN to perform entropy regularized policy update in closed form and investigate the stability of this approach in a deep RL context, which is the main contribution of this paper. Regarding the learning of the Q-function, we do not use LSTDQ \citep{lspi}. Instead, we implemented a more streamlined training procedure using a (neural) fitted Q-iteration scheme, similar to DQN \citep{dqn}---although using the Bellman operator of the current policy instead of the Bellman optimality operator in DQN---for simultaneously training the features and learning the Q-function of the current policy. However, we discuss how Cascade-NN techniques presented here can be used in the context of our work at the end of Section~\ref{sec.method}.


\section{The \ourAlgo algorithm}
\label{sec.method}
\begin{algorithm}[t]
   \caption{Pseudo code of the \ourAlgo Algortihm}
   \label{alg.micarl}

\begin{algorithmic}
   \STATE Set $\approxQ[{\policy[0]}]$ to the zero function
   \FOR{Iteration i in $\{1,\dots,\nbiter{}\}$}
	  \STATE Collect \nbsamp{} transitions of type $(s, a, r, s', a')$ from the environment following policy $\policy[i](s)\propto \exp \left( \eta \sum_{k=0}^{i-1} \approxQ[{\policy[k]}] (s, \cdot) \right)$
	  \STATE Add $n$ neurons to the current Cascade-NN
	  \STATE Set $\delta^0$ to the zero function
	  \FOR{Epoch e in $\{1,\dots,E\}$}
	    \STATE Compute target $r + \gamma (\approxQ[{\policy[k-1]}]+\delta^{e-1}) (s', a') - \approxQ[{\policy[k-1]}](s,a)$ for every transition $(s,a,r,s',a')$
	    \STATE Obtain $\delta^e$ by fitting the target using stochastic gradient descent
	  \ENDFOR 
	  \STATE Set $\approxQ[{\policy[k]}] = \approxQ[{\policy[k-1]}] + \delta^E$ 
   \ENDFOR
\end{algorithmic}
\end{algorithm}

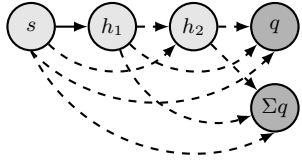
\begin{figure}
    \centering





\tikzstyle{output} = [circle, thick, minimum size = 0.7cm, draw = black, fill = black!30, 
                              inner sep=0pt, font = \small, node distance = 1.2cm]
\tikzstyle{other} = [circle, thick, minimum size = 0.7cm, draw = black, fill = black!10, 
                              inner sep = 0pt, font = \small, node distance = 1.2cm]
\tikzstyle{Tedge} = [->, dashed]
\tikzstyle{TCedge} = [->, dashed, bend right = 45]
\tikzstyle{Fedge} = [->, solid]
\tikzstyle{FCedge} = [->, solid, bend right = 45]

\begin{tikzpicture}[>=latex, thick, scale=1, every node/.style={scale=0.9}] \tiny
\node(s)[other, below of=i1]{$s$};
\node(h1)[other, right of=i2]{$h_{1}$};
\node(h2)[other, right of = h1]{$h_{2}$};
\node(o1)[output, right of =h2]{$q$};
\node(o2)[output, below of =o1]{$\Sigma q$};

\path    (s) edge[Fedge] (h1)
         (s) edge[TCedge] (h2)
         (h1) edge[Tedge] (h2)
         (s.south) edge[TCedge] (out.south)
         (h1) edge[TCedge] (out)
         (h2) edge[Tedge] (out)
         (s.south) edge[TCedge] (o2.south)
         (h1) edge[TCedge] (o2)
         (h2) edge[Tedge] (o2);

\end{tikzpicture}
    \caption{Our Q-function and policy network. The Cascade-NN has two heads, one storing the last Q-function---used to compute the targets of the neural fitted Q step in Alg.~\ref{alg.micarl}, while the output $\Sigma q$ accumulates all past weight matrices of node $q$, and is used by the softmax policy. Note that all nodes in the figure can be multi-dimensional, including hidden nodes/layers. See the implementation details for more information.}
    \label{fig.our_model}
\end{figure}

In this section, we introduce our algorithm \ourAlgo.
\ourAlgo{} is a deep RL algorithm that follows the policy iteration scheme. To approximate $Q_{\policy[k]}$, we use the adapted Cascade-NN model depicted in Fig.~\ref{fig.our_model}. This neural model has two heads giving the last Q-function and the sum of the last Q-functions. Alg.~\ref{alg.micarl} shows the pseudo-code for \ourAlgo. Starting with a zero function $\approxQ[{\policy[0]}]$, at every iteration $i>0$ we collect a dataset 
$\data[{\policy[i]}]$ of transition samples of type $(s, a, r, s', a')$, where actions are sampled from the current policy $\policy[i] \propto \exp (\eta \sum_{k = 0}^{i-1}\approxQ[{\policy[k]}])$ following the entropy regularized policy update of Eq.~\eqref{eq.policy}, and the next state $s'$ and rewards are given by the environment. Using the transition dataset $\data[{\policy[i]}]$, we learn the Q-function approximation $\approxQ[{\policy[k]}]$ following a standard (neural) fitted Q-iteration learning scheme. In our implementation, we learn $\delta_i = \Q[{\policy[i]}] - \Q[{\policy[i-1]}]$, instead of $\Q[{\policy[i]}]$. Starting with a zero function $\delta^0_i$, at each epoch $e$, we compute the targets $T^e_i(s, a)$ for every $(s, a, r, s', a') \in \data[{\policy[i]}]$,
\begin{equation}
\label{eq.target}
  T^e_i(s, a) = r + \gamma (\approxQ[{\policy[i-1]}]+\delta^{e-1}_i) (s', a') - \approxQ[{\policy[i-1]}](s,a).  
\end{equation}
We then update the learnable weights $\xi^{(i)}$
and $\deltaW[i]$ (see implementation details below) of the Cascade-NN for one epoch on the dataset $\data[{\policy[i]}]$ by minimizing the loss 
\begin{equation}
\label{eq.loss}
   \min_{\xi^{(i)}, \deltaW[i]}\sum_{(s, a,r,s',a') \sim \data[{\policy[i]}]} \left(\delta^e_i(s,a) - T^e_i(s, a)\right)^2,
\end{equation}
using stochastic gradient descent. Finally, the process of computing targets and fitting $\delta^e_i$ is repeated until reaching a given number of epochs $E$. 

As for the policy update, we simply add the new Q-function $\Q[{\policy[i]}] = \Q[{\policy[i-1]}] + \delta^E_i$ to the head of the Cascade-NN accumulating all past Q-functions, completing the definition of ${\policy[i+1]}$. Our implementation of \ourAlgo, including all experimental results, are obtained using an on-policy setting. We note however that the extension of our algorithm to the off-policy setting is trivial (one simply needs to replace $a'$ in a transition sample to match the current policy), but its investigation is left for future work.

\paragraph{Implementation details.}
Our Cascade-NN at any iteration $i$ is taking state $s \in \env$ as an input and outputs either a vector $\sum_{k=0}^{i-1} \approxQ[{\policy[k]}] (s, \cdot)$ or a vector $\approxQ[{\policy[i-1]}](s,\cdot)$. 
For a given Cascade-NN, we start with zero hidden neuron and we grow the architecture at each iteration $i\in\{1,\ldots, \nbiter{}\}$. In particular, at the beginning of iteration $i$ we have $(i-1)n $ hidden neurons and during the iteration we grow it by $n$ to reach a total of $i n $ neurons at the end of $i$-th iteration. Those hidden neurons are used to extract features. Further, what we consider as features is the direct inputs of output layers. At the beginning of the first iteration, the input 
is connected directly to the output layers, thus the input acts as the feature before the training.
 At any iteration, the input stays connected to the output layers and thus is a part of feature vectors.  
At the beginning of iteration $i$, its $(i-1)n$ old hidden neurons plus the input vector represent the feature vector function ${\feat[i-1]} : \env \rightarrow  \RR^{(i-1)n+\dim\env}$ that for any state $s\in \env$ returns its corresponding feature vector $\feat[i-1](s)$ from iteration $i-1$. Further, those neurons are frozen and not trained. Moreover, we have access to the two output layers described above. 
The weight matrix $\outW[i-1] \in \RR^{ |\act| \times \left((i-1)n+\dim\env\right)}$ corresponds to the sum of Q-values so that the matrix-vector product
$\outW[i-1] \feat[i-1] (s)$ gives the vector $ \sum_{k = 0}^{i-1}\approxQ[{\policy[k]}](s, \cdot)$ of size $| \act|$ (for simplicity, we omit bias term). 
Therefore, this output layer can be used to compute values necessary for policy $\policy[i]$ from Eq.~\eqref{eq.policy}. 
The weight matrix $\qW[i-1] \in \RR^{|\act| \times \left((i-1)n + \dim\env\right)}$ corresponds to the approximation Q-value of policy $\policy[i-1]$, that is $\approxQ[{\policy[i-1]}](s, \cdot) = \qW[i-1] \feat[i-1](s)$. 
During iteration $i$, $n$ new neurons are generated in a cascade manner as described in Section~\ref{sec.cascade}. They are responsible for computing a new component of the features, in particular they correspond to a vector function $\prefeat[i]: \RR^{n(i-1)+\dim\env} \rightarrow \RR^n$ that takes features from the previous iteration $\feat[i-1](s)$ as an input and outputs $\prefeat[i](\feat[i-1](s))$, which is further concatenated\footnote{We use $\concat$ to denote concatenation with respect to the last mode of a tensor, so in case of matrices $A \in \RR^{n \times k}$, $B\in \RR^{n \times m}$, then $C = \concat(A, B) \in \RR^{n\times(k+m)}$ is the matrix with first $k$ columns from $A$ and last $m$ columns from $B$.}
with $\feat[i-1](s)$ to constitute $\feat[i](s) = \concat\left(\feat[i-1](s), \prefeat[i](\feat[i-1](s))\right) \in \RR^{ni + \dim\env}$.
 Moreover, the new output layer is initialized with a weight matrix $\deltaW[i] \in \RR^{|\act| \times (ni+\dim\env) }$ that should represent $\delta_i(s, \cdot)=\approxQ[{\policy[i]}](s, \cdot) - \approxQ[{\policy[i-1]}](s, \cdot) = \deltaW[i] \feat[i](s)$. Parameters $\xi^{(i)}$ and $\deltaW[i]$ are optimized to minimize the loss defined in Eq.~\eqref{eq.loss}.
At the end of iteration $i$, once $\xi^{(i)}$ and $\deltaW[i]$ are optimized, $\approxQ[{\policy[i]}]$ can be evaluated with $\approxQ[{\policy[i-1]}](s, \cdot) + \delta_i (s, \cdot) = \qW[i-1] \feat[i-1](s) + \deltaW[i] \feat[i](s)$, therefore by setting $\qW[i] = \concat(\qW[i-1], O^{|\act| \times n}) + \deltaW[i]$, where  $O^{|\act| \times n}$ is a zero matrix of dimension $|\act| \times n$, the approximation $\approxQ[{\policy[i]}](s, \cdot)$ is naturally obtained from $\qW[i] \feat[i](s)$. Similarly, $\sum_{k=0}^{i} \approxQ[{\policy[k]}](s, \cdot) = \outW[i] \feat[i](s)$
where $\outW[i] = \concat(\outW[i- 1], O^{|\act| \times n}) + \qW[i]$.

In our implementation, the total number of weights grows as $\mathcal{O}(i^2)$ since new neurons connect to all previous neurons. This might be too prohibitive in practice, and we will experiment in future work with variants where new neurons only connect to a fixed number of past neurons. 
For example, following the correlation ideas of the original Cascade-NN (see Section~\ref{sec.cascade}), one might select the most promising past neurons according to the correlation between their activation and the current Bellman residual.

\section{Experiments}
\label{sec.exp}

\begin{figure}[t]
     \begin{subfigure}[b]{0.33\textwidth}
         \includegraphics[width=\textwidth]{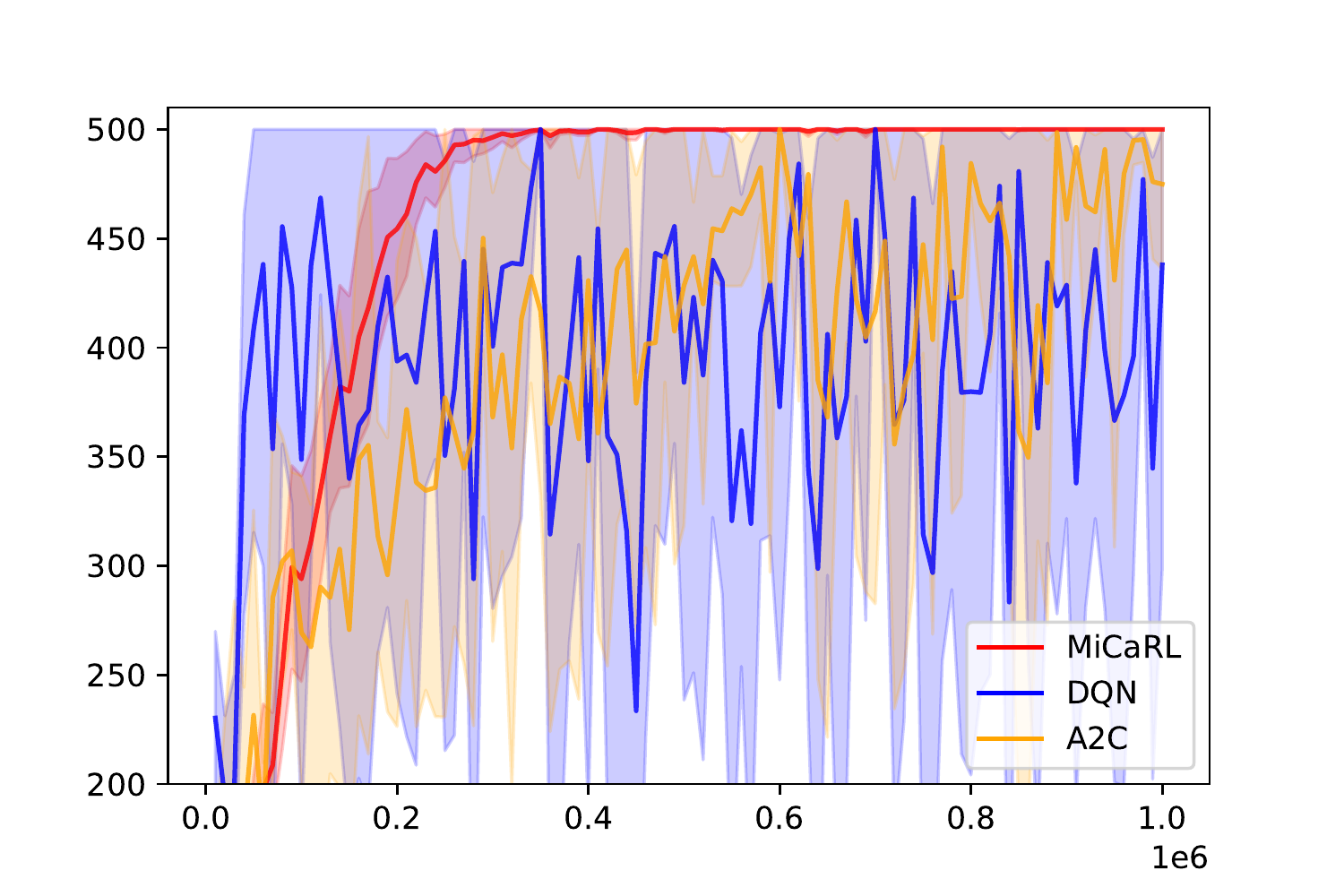}
     \end{subfigure}
     \begin{subfigure}[b]{0.33\textwidth}
         \includegraphics[width=\textwidth]{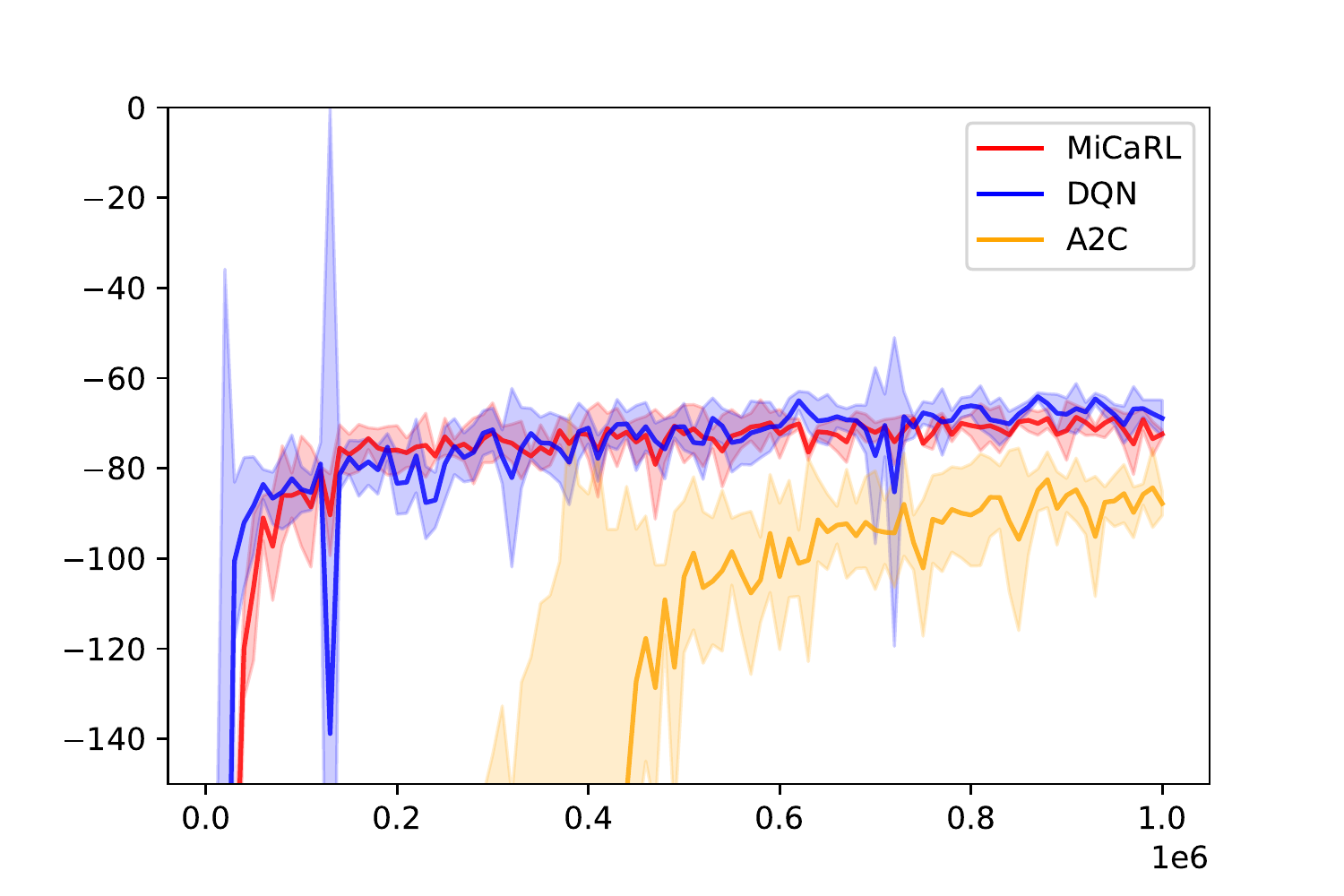}
     \end{subfigure}
     \begin{subfigure}[b]{0.33\textwidth}
         \includegraphics[width=\textwidth]{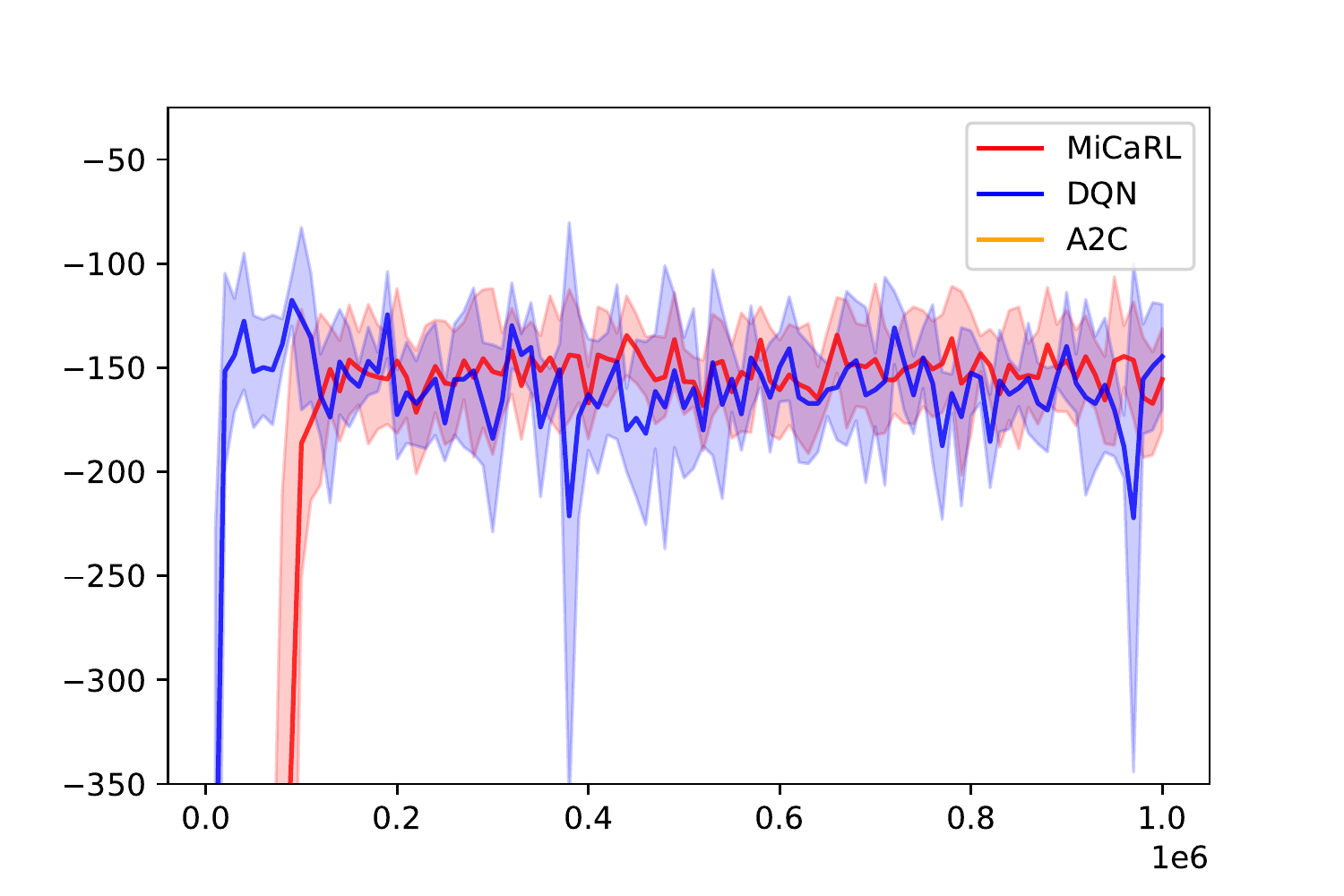}
     \end{subfigure}
     \caption{Zoom of the reward curves around their asymptotic values  on CartPole-v1 (left), Acrobot-v1 (center), and Pendulum-v1 (right).}\label{fig:zoom}
\end{figure}

\begin{figure}[t]
\centering
     \begin{subfigure}[b]{0.49\textwidth}
         \centering
         \includegraphics[width=\textwidth]{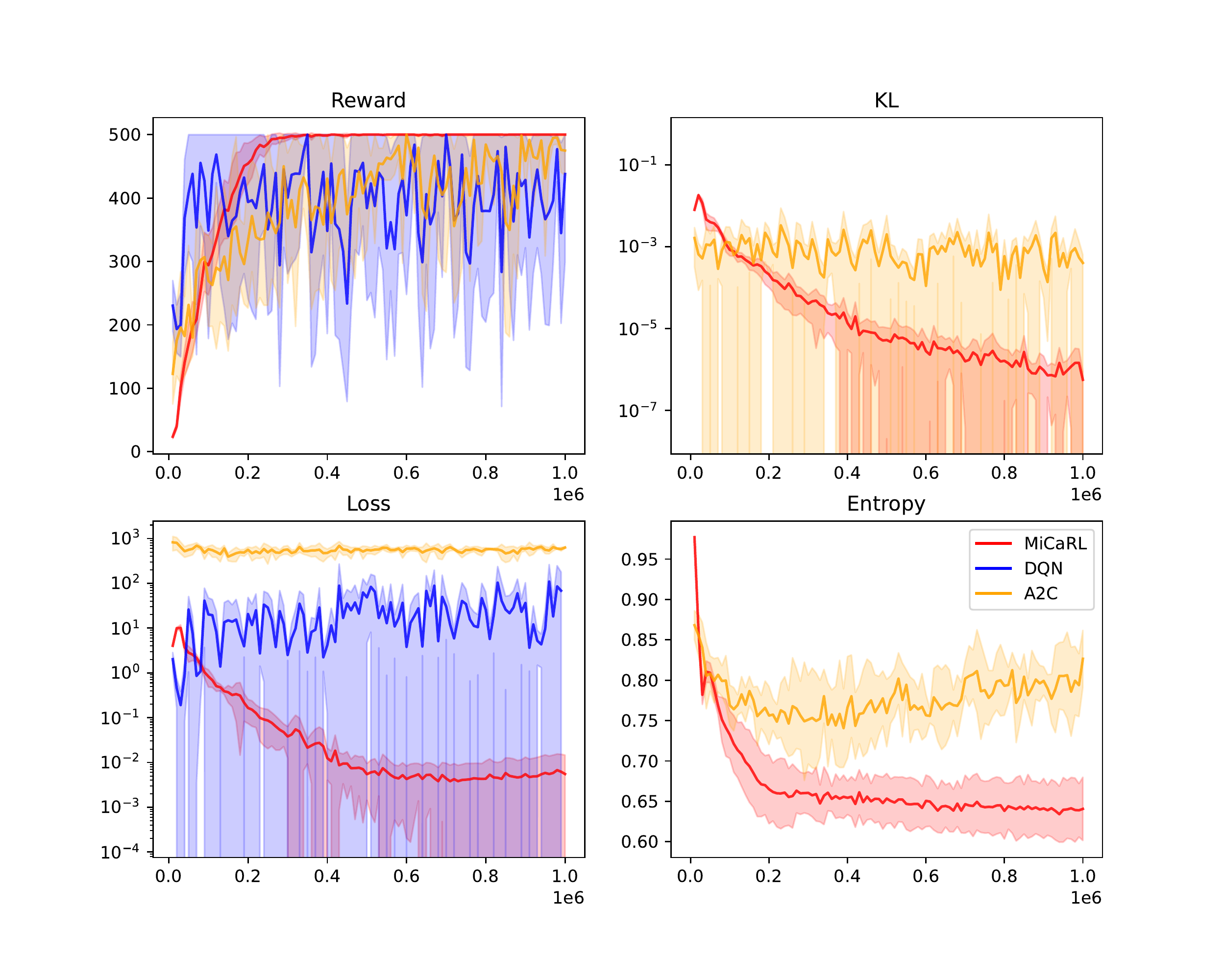}
     \end{subfigure}
     \begin{subfigure}[b]{0.49\textwidth}
         \centering
         \includegraphics[width=\textwidth]{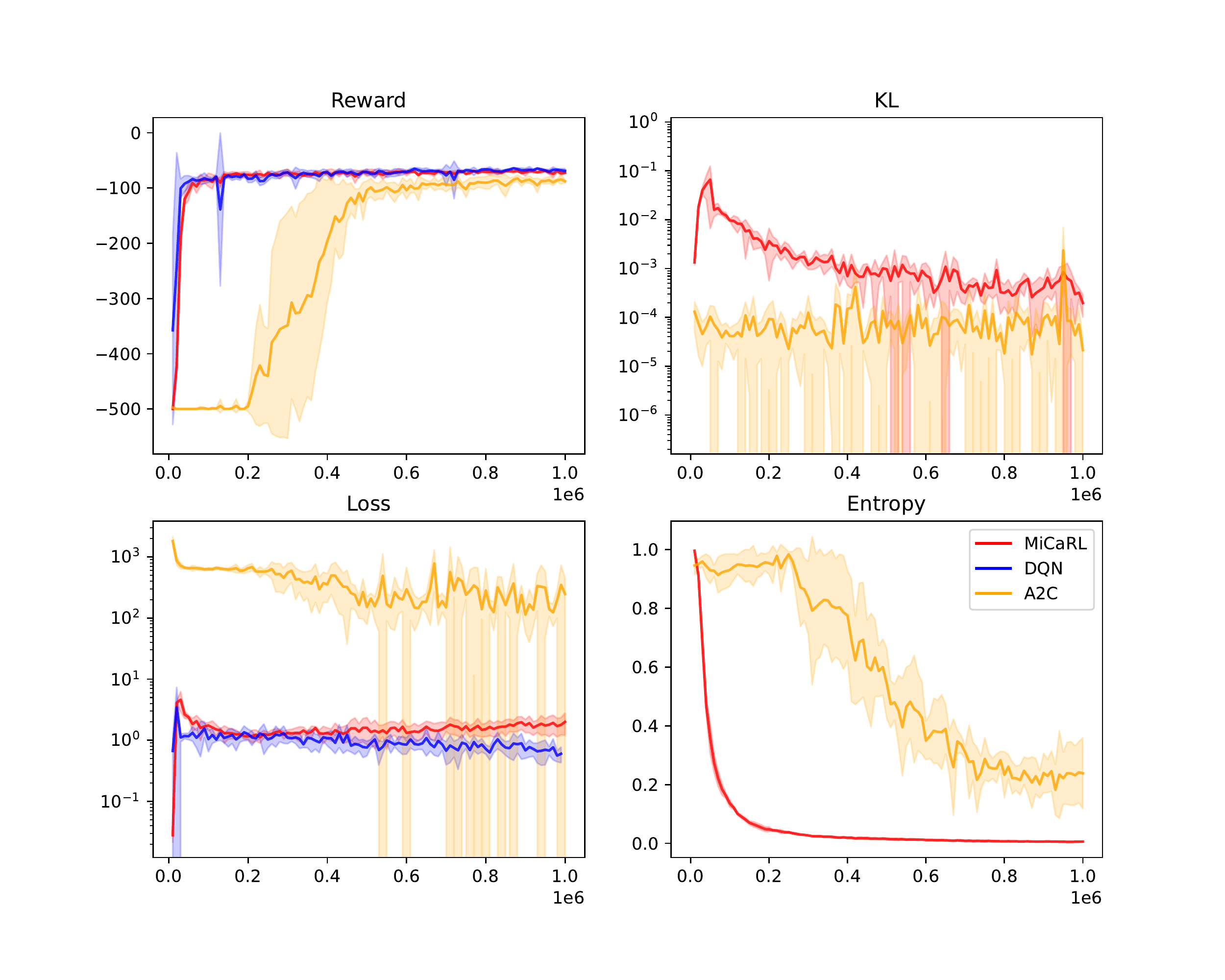}
     \end{subfigure}
     \begin{subfigure}[b]{0.49\textwidth}
         \centering
         \includegraphics[width=\textwidth]{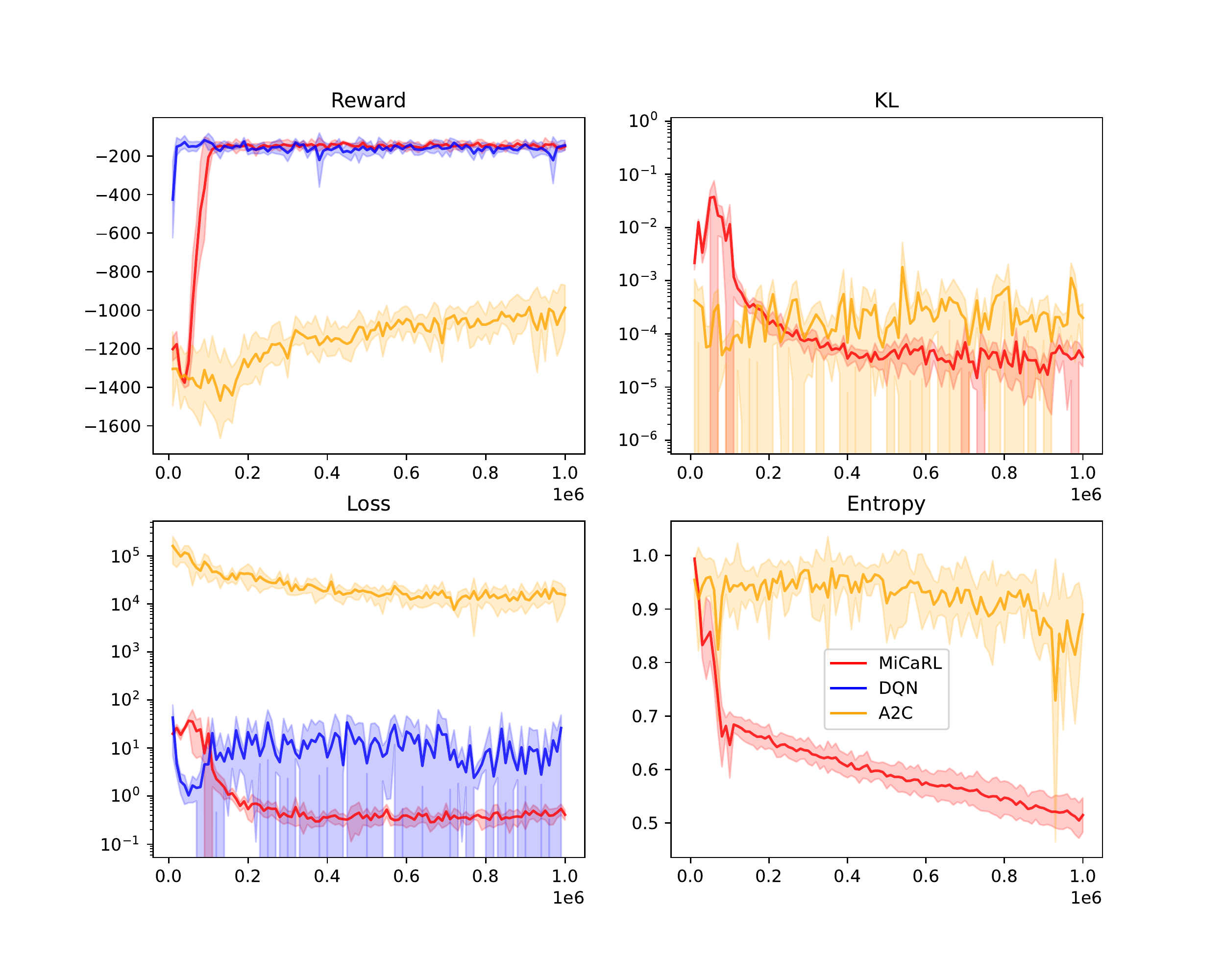}
     \end{subfigure}
     \begin{subfigure}[b]{0.49\textwidth}
         \centering
         \includegraphics[width=\textwidth]{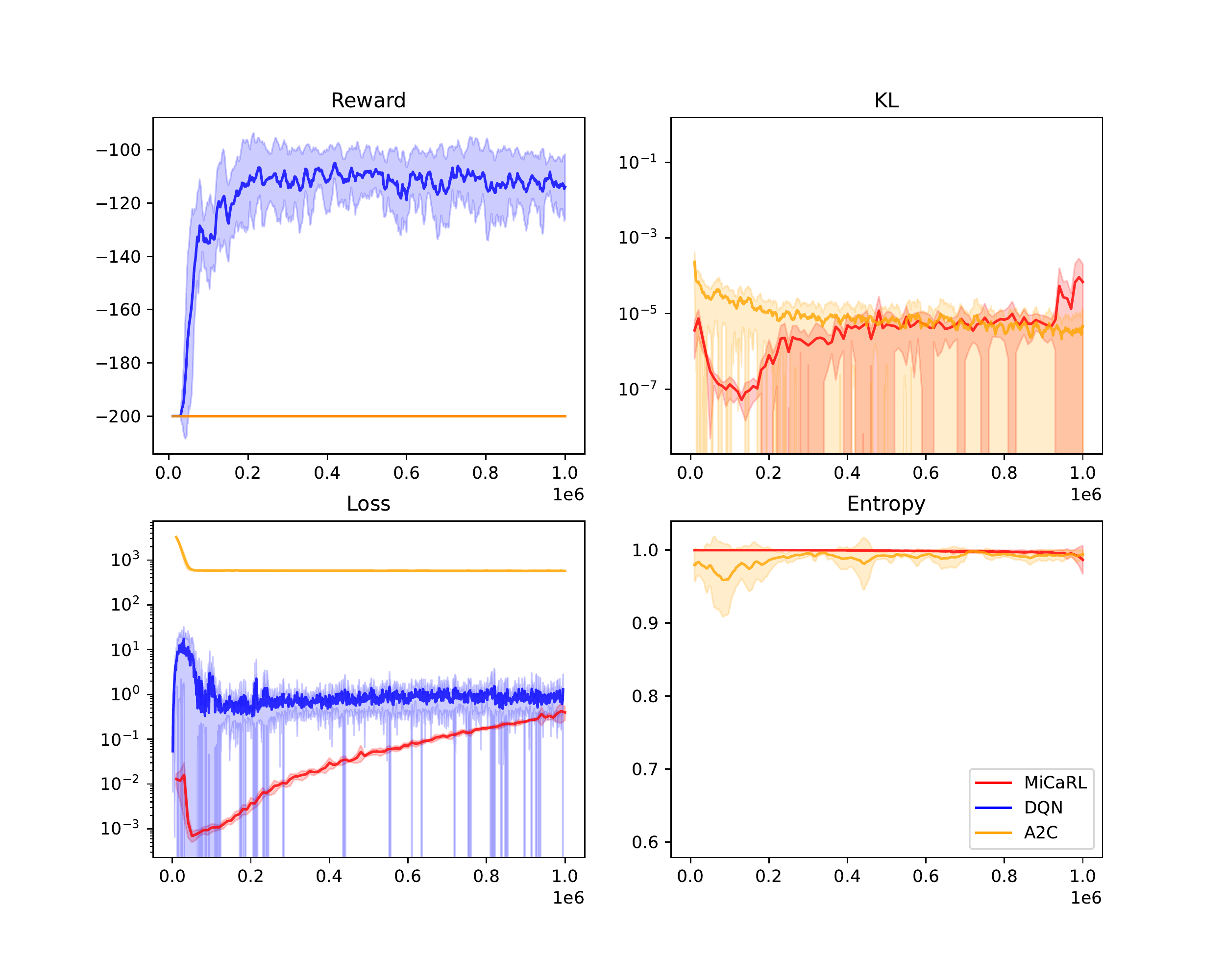}
     \end{subfigure}
     \caption{\ourAlgo{} and baselines on CartPole-v1 (top-left corner), Acrobot-v1 (top-right corner), Pendulum-v1 (bottom-left corner) and MountainCar-v0 (bottom-right corner).}\label{fig:4env}
\end{figure}

We evaluate \ourAlgo on four different gym environments: CartPole-v1, Acrobot-v1, a discrete action-space version of Pendulum-v1 and MountainCar-v0 and we compare it with A2C and DQN agents as implemented in \rlberry{} \cite{rlberry}.
In our experiments we use the default implementations of the agents: for the value function and the policy network in A2C, and for the Q-value function network of DQN, we take multilayer percecptrons with two hidden layers and 64 hidden units each.  
Our experiments show that \ourAlgo is in general more stable than both A2C and DQN and is achieving performances superior or comparable to DQN on three out of four environments, as it is evident from Figures~\ref{fig:zoom} and~\ref{fig:4env}. 
We plot curves averaged over five different seeds, using the default seeding handler from \citep{rlberry}, and the shaded area indicates one standard deviation. Figure~\ref{fig:zoom} is a zoom of the reward curves near the asymptotic values reached by the best performing algorithms. For this reason, for Pendulum-v1 we do not see the curve corresponding to A2C, since its performance is quite poor on this environment. In Figure~\ref{fig:4env}, we plot a more detailed study in which we do not just show rewards but also losses during training, average normalized entropy (across visited states) and the KL-divergence between two consecutive policies for A2C and \ourAlgo{} (not for DQN since it is not a policy iteration algorithm). Note that the way the loss is computed differs from one algorithm to another: \ourAlgo and DQN compute Bellman residuals of different Bellman operators, while A2C tries to minimize the difference between critic output and Monte Carlo estimate of the value function based on the dataset. This explains different levels of loss values in loss plots.

We note that the number of neurons $n$ to add at every iteration is important. For small $n$, we typically lose the good properties of over-parameterization when learning $\delta$. Especially, since in our implementation we do not use the Cascade-NN idea discussed in Section~\ref{sec.cascade} of training a number of candidate neurons $>n$ before picking the $n$ most promising ones. As a result, we found that using small $n$, e.g. $n = 1$, would often lead to dead neurons and more generally, to attraction to poor local optima. We address this issue by using a larger than usual $n$, typically $n\in\{10, 20, 50\}$.
We try also combinations of the other hyper-parameters: the strength parameter of the KL-regularizer in Eq.~\eqref{eq.mirror} is $\eta\in \{0.01, 0.1, 0.5, 1, 5\}$ and the number of epochs per iteration $E\in \{ 64, 128\}$. Hyper-parameter search is done in grid search manner with the help of Ray Tune software~\cite{raytune}.
Results are quite similar for different choices of epochs, therefore we further keep it fixed $E=64$. In contrast, our algorithm is more sensitive to the choice of $\eta$ and slightly less to the choice of $n$. 
Further, we report separately for each environment the best choice of $n$ and $\eta$.
In addition, each trial is conducted with a number of iterations equal to $\nbiter=100$, $\bsize=64$ and in each iteration we take $\nbsamp=10\,000$ steps in the environment. Therefore, the total number of steps in the environment is equal to $1\,000\,000$ for each experiment, which is the quantity on the x-axis in Figures \ref{fig:4env} and \ref{fig:zoom}. The final architecture of \ourAlgo contains $\nbiter\times n$ hidden neurons.

The strong stability of \ourAlgo{} is probably due to the second term in Eq.~\eqref{eq.mirror} that acts as a regularizer. Nonetheless, \ourAlgo{} achieves impressive and surprising performances. For example on CartPole-v1, \ourAlgo{} is able to find a much better approximation of the Q-value function with respect to our baselines and the reward remains stable at its maximum value after a brief training phase. 
On Acrobot-v1 and Pendulum-v1, \ourAlgo is performing as good, or better, than DQN, and significantly better than A2C. 
MountainCar-v0 is instead known to be a difficult environment and \ourAlgo is not able to learn anything, the same for A2C. In contrast, DQN has a good performance on this environment as well.

\paragraph{CartPole-v1}
We test the performance of \ourAlgo{} against the baselines of A2C and DQN on CartPole-v1 in Figure~\ref{fig:4env}. The best performance of \ourAlgo is observed for $n=10$ and $\eta=0.1$.
The training loss of \ourAlgo is getting closer to zero than the losses of the two other algorithms, showing that \ourAlgo succeeds in better approximating Q-value functions. Furthermore, A2C and DQN loss curves are plateauing and oscillating, 
proving them to be unstable. 
The curves of the rewards for A2C and DQN also exhibit an oscillating behaviour. Sometimes they manage to reach the optimum value, which is 500 for this environment, but on average they do not manage to maintain this performance in all episodes. The stability of \ourAlgo is also confirmed by the KL-divergence plot where at the end of the training the difference between successive policies becomes increasingly negligible, meaning that the algorithm is converging to a policy. 
Interestingly, the normalized entropy (having value between 0 for a deterministic policy and 1 for the uniform policy) of the policy is staying quite high, around $0.65$, for  \ourAlgo{}.
This reflects the fact that under a near optimal policy that balances the pole correctly, there are a lot of states where both actions are viable, meaning that our algorithm not only finds an optimal policy, but several optimal ones. A2C exhibits an even higher entropy, but since its reward is still oscillating, one cannot draw the same conclusions.
Overall, \ourAlgo achieves the best results, while being more stable than other baselines.

\paragraph{Acrobot-v1 and Pendulum-v1}
In the case of Acrobot-v1, \ourAlgo performs the best with $n=50$, $\eta=1$, while in case of Pendulum-v1 with $n=50$ and $\eta=0.1$. Results on Acrobot-v1 and a discrete version of Pendulum-v1 are almost identical to DQN in terms of the rewards and both significantly outperform A2C. We note that due to the intrinsic stability of \ourAlgo{}, it shows even less oscillating behaviour.
The loss curves of \ourAlgo{} and DQN are very similar for Acrobot-v1, while, we note that for Pendulum-v1 \ourAlgo{} is fitting much better than DQN the loss function. 

\paragraph{MountainCar-v0}
We show the performance of \ourAlgo on MountainCar-v0 with $n=50$ and $\eta=0.1$, but the results for all the other hyperparameters are very similar.
Due to its very sparse reward, MountainCar-v0 is usually a very hard task for reinforcement learning agents. In this case, the normalized entropy of the policy is very close to one. This indicates that the policy is sampling actions from the uniform distribution, which seemingly prevents the discovery of any postive reward. As all rewards that the agent observes are $-1$, the learned Q-function is a completely flat function, which according to Eq.~\eqref{eq.policy} results in a uniform distribution, explaining why the entropy stays close to one. All in all, while entropy regularization is a good heuristic for maintaining high entropy and sustaining exploration, we can see that it does not fully address the exploration problem in RL. 

\paragraph{Summary of results.}
To summarize, we can categorize the results in three. In i) the best case scenario, on CartPole-v1, we observed formidable convergence on all metrics: the cumulative reward reaching and staying at its maximal value, the mean squared Bellman residual going to $10^{-3}$, and the KL-divergence between successive policies reaching zero. In ii) the middle case on Pendulum-v1 and Acrobot-v1, while results are on par with the state-of-the-art with perhaps an ever so slightly higher stability of the cumulative rewards, there still remains a persistent Bellman error despite the ever growing size of the Cascade-NN and an abundance of data on relatively simple problems. This is somehow unsatisfying and indicates that research is still needed to discover better policy evaluation algorithms. Finally, in iii) the failure case on MountainCar-v0, where the policies at all iterations are close to uniform over the action space, suggesting that while entropy regularization is a good exploration mechanism with theoretical convergence guarantees, in practice, other exploration methods might still be necessary in some more challenging sparse reward cases.

\section{Related Work}
\label{sec.rel}
Examples of deep RL algorithms implementing entropy regularization include TRPO \citep{Schulman15}, SAC \citep{Haarnoja18} and MPO \citep{Abdolmaleki18}, but the closest to our work is \politex~\citep{politex, improvedpolitex}. Similarly to \ourAlgo{}, in \politex, the new policy is defined as a Boltzmann distribution
over the sum of all past state-action value estimates, resulting from a KL-divergence regularization on the policy update, which makes the learning 
process less noisy.
For example, the experimental results of~\citep{improvedpolitex} show good convergence results,
outperforming other policy optimization algorithms. However, to perform this policy update in closed form,  \cite{politex} considered learning separate
state-action value networks which is a computationally expensive process requiring
to store all past NNs corresponding to different Q-functions. The more
recent implementation of~\citep{improvedpolitex} avoids keeping different
NNs and instead relies on one NN and experience replay buffer to approximate 
directly the average behaviour of all previous state-action value functions.



The strategy of \politex requires the knowledge of all previously trained state-action value functions. Implementing them with NNs is challenging as the candidate NN should be able to approximate equally good all the old functions together with the new one. This particular problem known as catastrophic forgetting is studied in the field of Incremental Learning. 
Among the approaches of Incremental Learning, one could distinguish three main directions: architectural, regularization and rehearsal strategies. Architectural strategies, as used in \ourAlgo, suggest modifying (e.g. expanding) the NN structure in order to preserve the good performance on the old tasks and achieve good precision on the new one~\citep{progressiveNN,cascadeNN}. Regularization techniques may be divided in two groups. The first group is weight regularization, which is typically done by introducing the additional component in the loss whose goal is  to penalize the change of the weights that are important for the old tasks (e.g.~\citep{ILweightReg,MAS}). Second group is knowledge distillation~\citep{lwf,GDKD}, which was firstly used for transfer learning, but can be efficiently applied to Incremental Learning, by bringing the "knowledge" from the old model to the new one, forcing the output of the new model to be more consistent with the output of old models. Finally, rehearsal methods~\citep{iCarl} or pseudo-rehearsal methods~\citep{sigann} alleviate the problem of catastrophic forgetting by reusing the data of the old tasks when learning the new one. Pseudo-rehearsal methods slightly differ from rehearsal methods as they generate data from the old tasks instead of storing it. More thorough overview of Incremental Learning methods can be found in~\citep{ICsurvey}.




\section{Conclusion}
\label{sec.conclusion}

In this paper, we consider entropy regularized reinforcement learning algorithms and how to implement them efficiently in practice. These algorithms were shown to be more stable than the classical approaches as they constrain the current policy to be closer to the previous ones. In our study, we concentrate on the \politex algorithm that builds the policy from the output of all  Q-value functions of past policies. Implementing those functions in practice is not trivial, and straightforward approaches of assigning one neural network per Q-value function is only possible for very small tasks and cannot scale. Instead, we suggest using \ourAlgo based on Cascade-NN, a neural architecture that grows each iteration by $n$ neurons, capable of preserving past information while new neurons can leverage increasingly richer feature representations. Our preliminary results show that this novel approach can successfully compete with the state-of-the-art for most of the considered use cases, while exhibiting impressive convergence on all considered metrics for CartPole-v1. We believe these are very promising preliminary results suggesting that \ourAlgo and its non-parametric approach to Q-function approximation is worth further investigation. In future work, we would like to understand what makes the algorithm better approximate the Q-function on CartPole-v1 than the other problems, and research the integration of more sophisticated exploration mechanisms in \ourAlgo to tackle sparse reward problems. 

\acks{}
The authors would like to acknowledge the financial support of the French Ministry of Higher Education and Research, Inria, the Hauts-de-France region. Philippe Preux is also supported by the Métropole Européenne de Lille, through the AI chair Apprenf number R-PILOTE-19-004-APPRENF.
Riccardo Della Vecchia is thankful for the funding received by the CHIST-ERA Project Causal eXplainations in Reinforcement Learning -- CausalXRL. \footnote{\url{https://www.chistera.eu/projects/causalxrl}}
Alena Shilova acknowledges the funding coming by the Challenge HPC-BigData INRIA Project LAB. \footnote{\url{https://project.inria.fr/hpcbigdata/}}
We also thank the Scool team at Inria Lille Nord Europe.

\vskip 0.2in
\bibliography{sample}


\newpage

\appendix

\section{Number of added neurons}

\begin{figure}[th]
\centering
     \begin{subfigure}[b]{0.47\textwidth}
         \includegraphics[width=\textwidth]{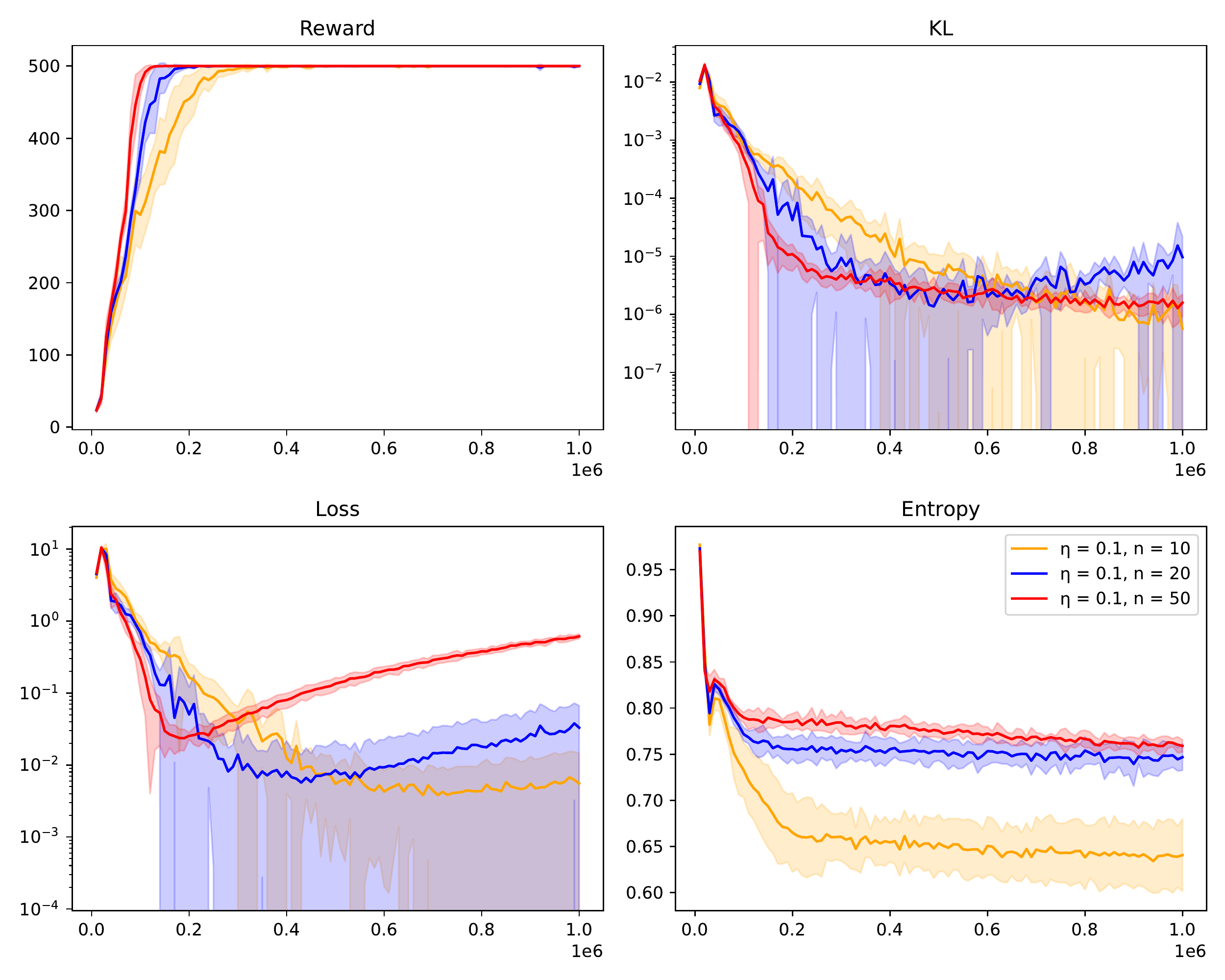}
     \end{subfigure}
     \hfill
     \begin{subfigure}[b]{0.47\textwidth}
         \includegraphics[width=\textwidth]{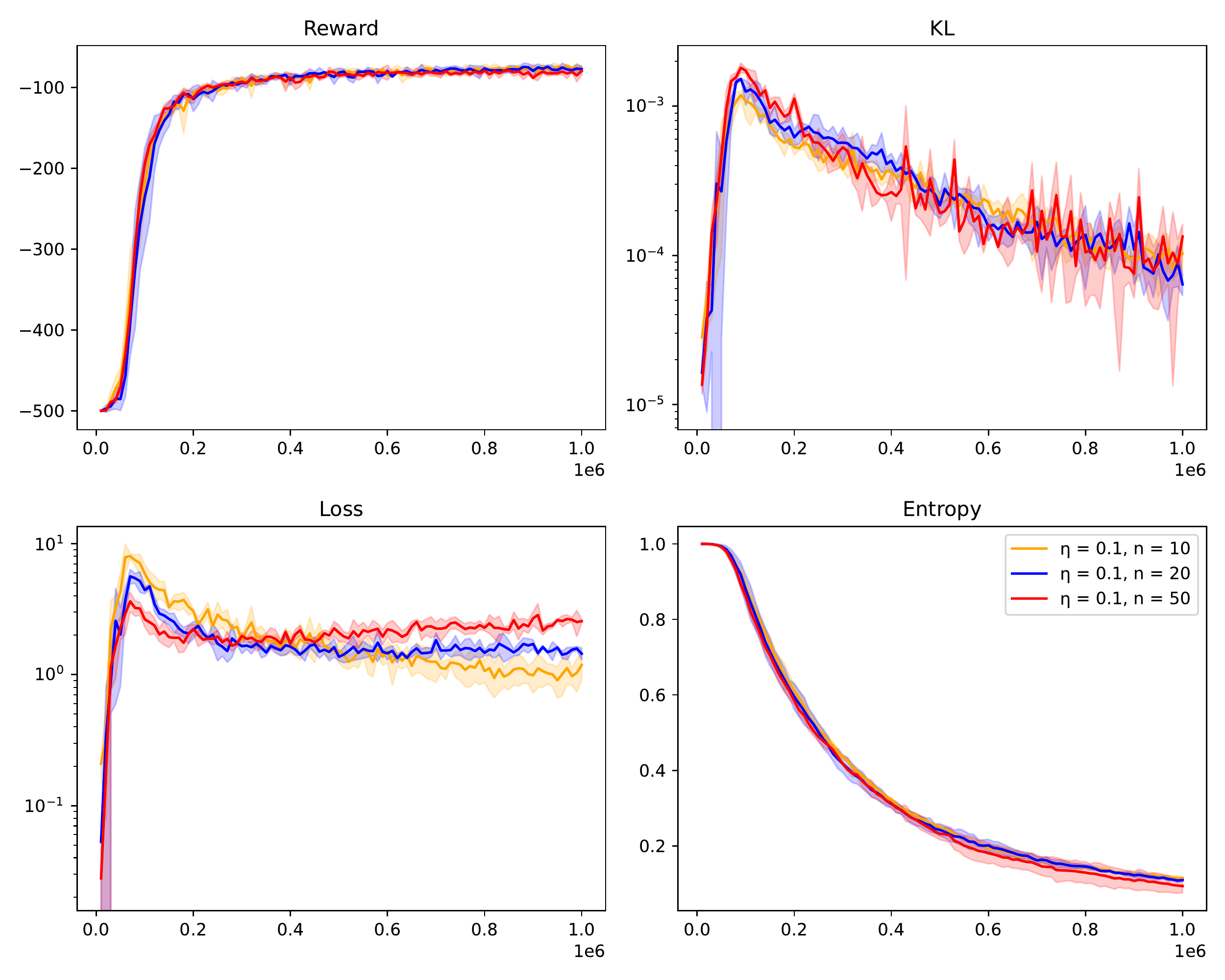}
     \end{subfigure}
     \begin{subfigure}[b]{0.47\textwidth}
         \includegraphics[width=\textwidth]{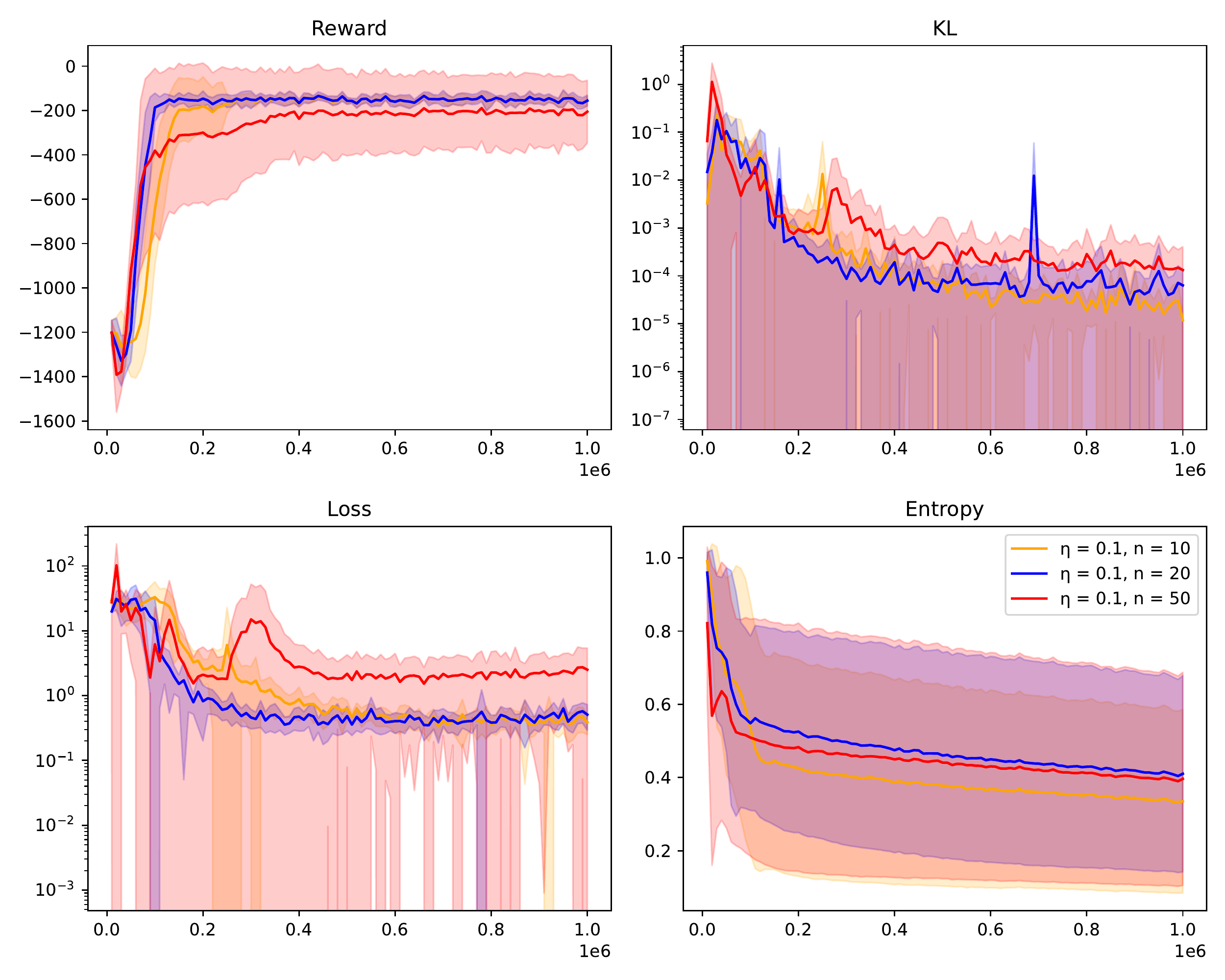}
     \end{subfigure}
     \hfill
     \begin{subfigure}[b]{0.47\textwidth}
         \includegraphics[width=\textwidth]{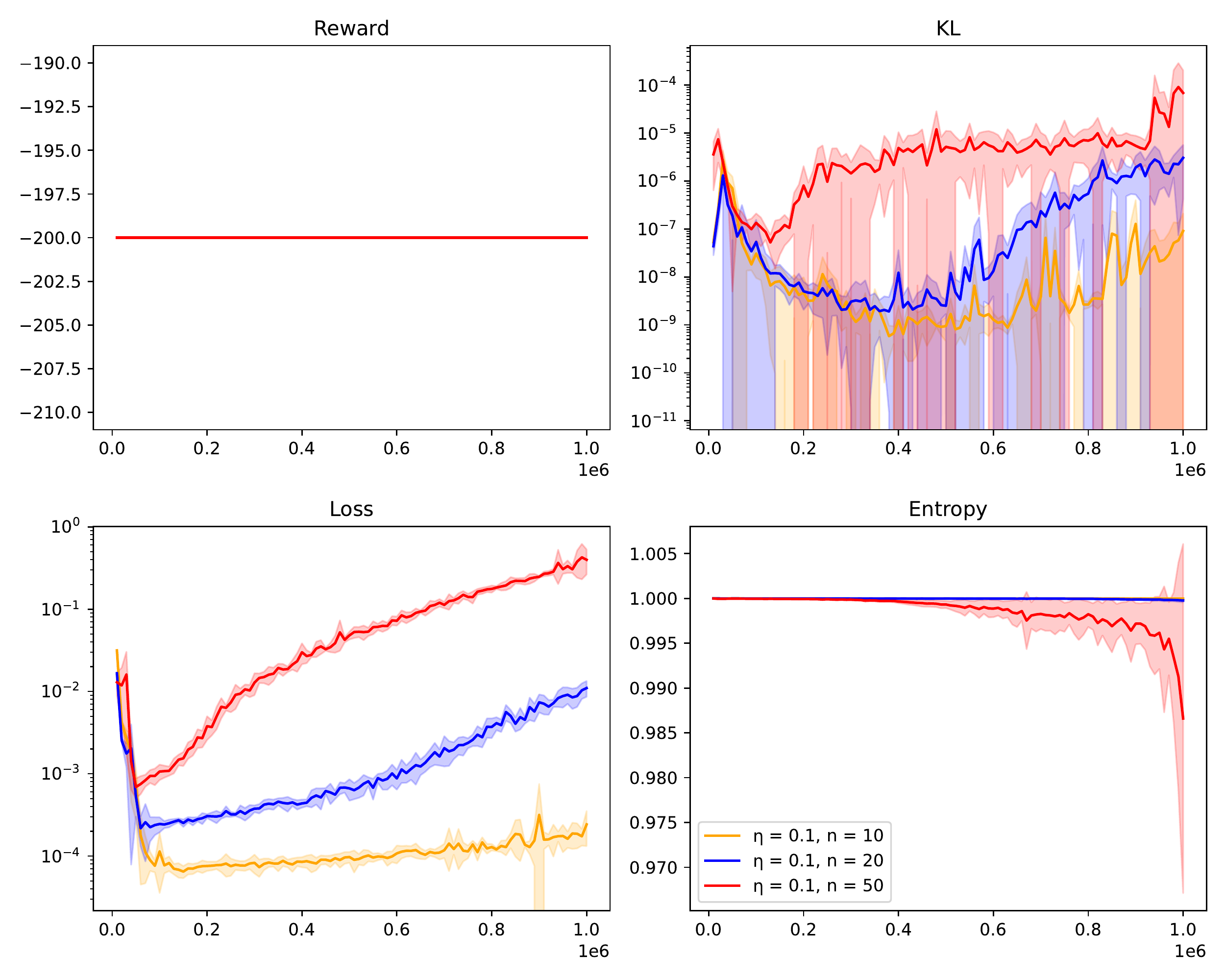}
     \end{subfigure}
     \caption{Dependence on the number of added neurons per iteration $n$ on CartPole-v1 (top-left corner), Acrobot-v1 (top-right corner), Pendulum-v1 (bottom-left corner) and MountainCar-v0 (bottom-right corner).}\label{fig:different_n}
\end{figure}

We ran an additional set of experiments to study the performance of \ourAlgo{} with respect to its dependence on $n$, the number of neurons added per iteration, in Figure \ref{fig:different_n}. In general, we expect better performance as we increase the number of neurons since it should be easier in this case to approximate the Q function. This is confirmed by our simulations for $n\in\{10,20,50\}$, where we see that more neurons seem to indicate faster learning, as least initially, while in some cases, the mean squared Bellman residual seem to increase with time for higher $n$, perhaps due to an overfitting problem. Asymptotically however, the curves of the rewards achieve approximately the same values.

\end{document}